\documentclass{preprint}

\usepackage{amsmath}
\usepackage{xcolor}
\usepackage{booktabs}
\usepackage{graphicx} 
\usepackage[numbers,comma,sort&compress]{natbib}
\usepackage[colorlinks=true]{hyperref}
\usepackage{multirow}
\usepackage{xspace}
\usepackage{array}

\title{Uncovering Misattributed Suicide Causes through Annotation Inconsistency Detection in Death Investigation Notes}
\author[1]{Song Wang}
\author[2]{Yiliang Zhou}
\author[3]{Ziqiang Han}
\author[4]{Cui Tao}
\author[2]{Yunyu Xiao}
\author[5]{Ying Ding}
\author[1]{Joydeep Ghosh}
\author[2,*]{Yifan Peng}
\affil[1]{Cockrell School of Engineering, The University of Texas at Austin, Austin, TX, USA}
\affil[2]{Population Health Science, Weill Cornell Medicine, New York, NY, USA
}
\affil[3]{School of Political Science and Public Administration, Shandong University, Qingdao, Shandong, China}
\affil[4]{Department of Artificial Intelligence and Informatics, Mayo Clinic, Rochester, Minnesota, USA}
\affil[5]{School of Information, The University of Texas at Austin, Austin, TX, USA}
\affil[*]{Corresponding: \url{yip4002@med.cornell.edu}}

\newcommand{\fscore}{\text{F1}\xspace}

\begin{document}

\maketitle

\begin{abstract}
Data accuracy is essential for scientific research and policy development. The National Violent Death Reporting System (NVDRS) data is widely used for discovering the patterns and causes of death. Recent studies suggested the annotation inconsistencies within the NVDRS and the potential impact on erroneous suicide-cause attributions. We present an empirical Natural Language Processing (NLP) approach to detect annotation inconsistencies and adopt a cross-validation-like paradigm to identify problematic instances. We analyzed 267,804 suicide death incidents between 2003 and 2020 from the NVDRS. Our results showed that incorporating the target state's data into training the suicide-crisis classifier brought an increase of 5.4\% to the \fscore score on the target state's test set and a decrease of 1.1\% on other states' test set. To conclude, we demonstrated the annotation inconsistencies in NVDRS's death investigation notes, identified problematic instances, evaluated the effectiveness of correcting problematic instances, and eventually proposed an NLP improvement solution.
\end{abstract}

\section{Introduction}\label{introduction}

In recent years, the United States (U.S.) has experienced a concerning increase in suicide deaths, marked by an alarming 36\% suicide rate rise between 2000 and 2021\footnote{\url{https://www.cdc.gov/suicide/facts/index.html}}. Understanding the suicide causes is critical and essential for effective interventions and suicide prevention policymaking.

The National Violent Death Reporting System (NVDRS) is a comprehensive surveillance initiative gathering violent fatality data from all 50 U.S. states, the District of Columbia, and Puerto Rico\footnote{\url{https://www.cdc.gov/violenceprevention/datasources/nvdrs/index.html}}. It meticulously documents information about suicide victims, including demographics and vital social determinants of health. The database also contains detailed death investigation notes for each incident, describing the circumstances potentially contributing to the suicide. The NVDRS coded a series of suicide circumstance variables, which were manually annotated by human abstractors utilizing the information contained in the death investigation notes. These suicide circumstance variables indicated the presence status of suicide-related social factors (e.g., Family Relationship Crisis, Mental Health Crisis, and Physical Health Crisis). To ensure data quality, the NVDRS offers a standardized coding manual for those working with the database. The program also conducts routine coding training for newly onboarded abstractors and provides continuous coding support. However, it is noteworthy that only 5\% of the incidents' annotations were verified by two independent annotators, leaving an overwhelming 95\% of the data reliant on a single annotator~\cite{Liu2023-ry}. The lack of a peer verification process increases the risk of annotation discrepancies between individual annotators, which could lead to state-level inconsistencies and even intra-state inconsistencies. Moreover, despite abstractors adhering to guidelines, there still exists a potential for annotation inconsistencies due to the possibility of inadequate expertise and human errors~\cite{Hollenstein2016-vw}.

In our prior research, we developed Natural Language Processing (NLP) methods to extract suicide circumstances from the NVDRS narratives~\cite{Wang2023-ov}. Our findings highlighted the performance disparities across states and identified the inconsistencies in the NVDRS data annotations. Several studies have tackled data annotation errors in NLP through various approaches~\cite{Kveton2002-nf,Ma2001-vx,Ule2004-ri,Loftsson2009-dp,Kato2010-gg,Manning2011-bq,Nguyen2015-ul,Zeng2021-st}, for example, utilizing conventional probabilistic approaches~\cite{Chong2022-fa}, training conventional machine learning models (e.g., Support Vector Machines)~\cite{Eskin2000-bn,Nakagawa2002-tw,Dligach2011-df,Amiri2018-hj,Swayamdipta2020-yr,Yaghoub-Zadeh-Fard2019-dd,Wang2019-zf,Northcutt2021-ee}, developing generative models via active learning~\cite{Rehbein2017-ld}, or utilizing pre-trained language models~\cite{Chong2022-fa}. However, the conventional probabilistic approaches cannot handle infrequent events or compare events with similar probabilities. This is primarily because the probabilities cannot be calculated or compared with high confidence. At the same time, the conventional supervised training paradigm needs high-quality annotated data during the training process. This poses a limitation when applying these methods to the NVDRS dataset, where only 5\% of the data were verified by two annotators. Moreover, previous attempts mainly focused on certain NLP tasks, such as Part-of-Speech (POS) tagging and Named Entity Recognition (NER). Those approaches cannot be directly applied to identifying mis-labelings in free-text death investigation notes.

This study introduced an empirical NLP approach utilizing transformer-based models to uncover data annotation inconsistencies in death investigation notes. In our evaluation, we measured the annotation discrepancies across all U.S. states. Here, we refer to the state under evaluation as the `target state' and all other states as the  `other states'. Our focus was on three specific suicide circumstance variables: Family Relationship Crisis, Mental Health Crisis, and Physical Health Crisis. These variables were selected for their higher prevalence of positive instances in the National Violent Death Reporting System (NVDRS) dataset, and their poor classification scores, as demonstrated in prior work~\cite{Wang2023-ov}. We calculated the annotation inconsistencies by determining the degree of decrease in the \fscore score when the model training data was switched from data sampled from the target state to data sourced from other states. We also designed a cross-validation-like framework to identify problematic data instances contributing to these inconsistencies. These instances were then manually rectified, and the corrected labels were re-evaluated. In this work, we used \fscore scores as an underlying evaluation metric for comparison. The \fscore score is the harmonic mean of Precision (the ratio of true positive predictions to all positive predictions) and Recall (the ratio of true positive predictions to all actual positives). The \fscore score balances Precision and Recall into one single value. A higher \fscore score indicates better model performance. Our experiments showed the efficacy of our approach in identifying potential annotation errors in NVDRS's death investigation notes. Moreover, correcting these errors yielded an average \fscore score improvement of 3.85\%. Finally, we analyzed the Odds Ratio (OR) computed for various demographic subgroups (age, sex, race) to better understand the risk of bias.

In summary, our work aims to enhance the current understanding of annotation inconsistencies in unstructured death investigation notes in the NVDRS. By addressing the inconsistencies, our work hopes to pave the way for more accurate and reliable utilization of NVDRS data in discovering suicide causes and developing suicide prevention strategies at the national, state, and local levels.

\section{Results}\label{results}
\subsection{Validating Annotation Inconsistency}\label{validating-annotation-inconsistency}

\begin{figure}[hbt!]
    \centering
    \includegraphics[width=.9\textwidth]{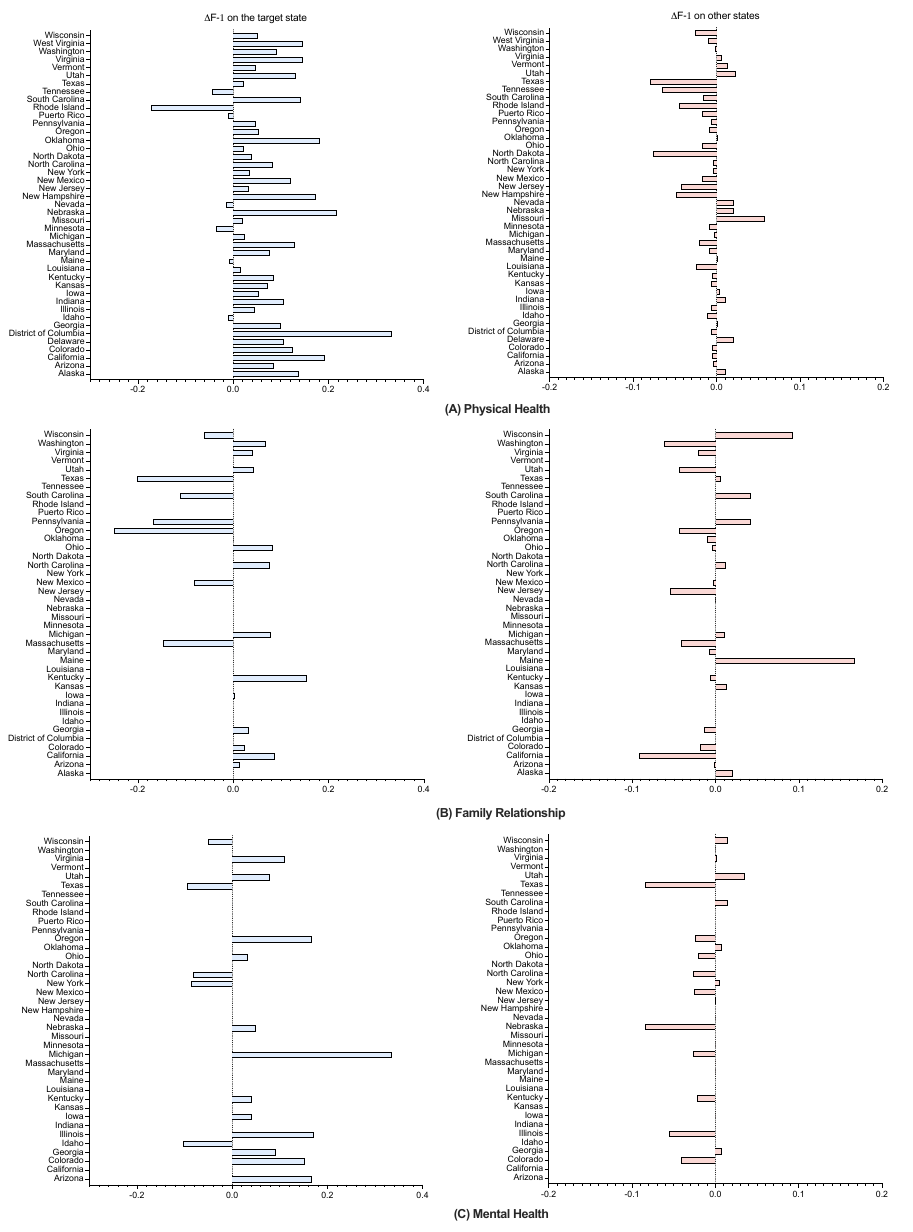}
    \caption{\(\Delta \fscore\)'s on the test sets of the target state (\(\uparrow\)) and other states (\(\downarrow\)). (A) Physical Health, (B) Family Relationship, (C) Mental Health Crisis.}
    \label{fig:figure1}
    \vspace{1em}
\end{figure}

For each crisis, states with fewer than 10 positive instances were excluded to ensure adequate training data for validating annotation inconsistency. In Figure~\ref{fig:figure1}(A), outcomes for the Physical Health Crisis show that when the target state's data was added to the training set, approximately 83.7\% (36 out of 43) of states improved their prediction performance on the target state's test set (indicated by a positive \(\Delta \fscore\)), while around 69.8\% (30 out of 43) of states experienced a performance drop on other states' test sets (indicated by a negative \(\Delta \fscore\)).

In Figure~\ref{fig:figure1}(B), results for the Family Relationship Crisis reveal that when the target state's data was included in the training, 32.5\% (13 out of 40) of states improved their prediction performance on the target state's test set. In comparison, 40\% (16 out of 40) of states experienced a performance decrease on other states' test sets.

In Figure~\ref{fig:figure1}(C), findings for the Mental Health Crisis demonstrate that after including the target state's data in training, approximately 33.3\% (13 out of 39) of states improved their prediction performance on the target state's test set. In comparison, around 43.6\% (17 out of 39) of states saw a performance drop on other states' test sets.

The performance variation with different training data combinations suggests annotation inconsistencies across states. These findings highlight the need to rectify label inconsistencies in death investigation notes to enhance data quality and better understand suicide causes.

\subsection{Discovering Problematic Instances}\label{discovering-problematic-instances}

\begin{figure}[hbt!]
    \centering
    \includegraphics[width=.55\textwidth]{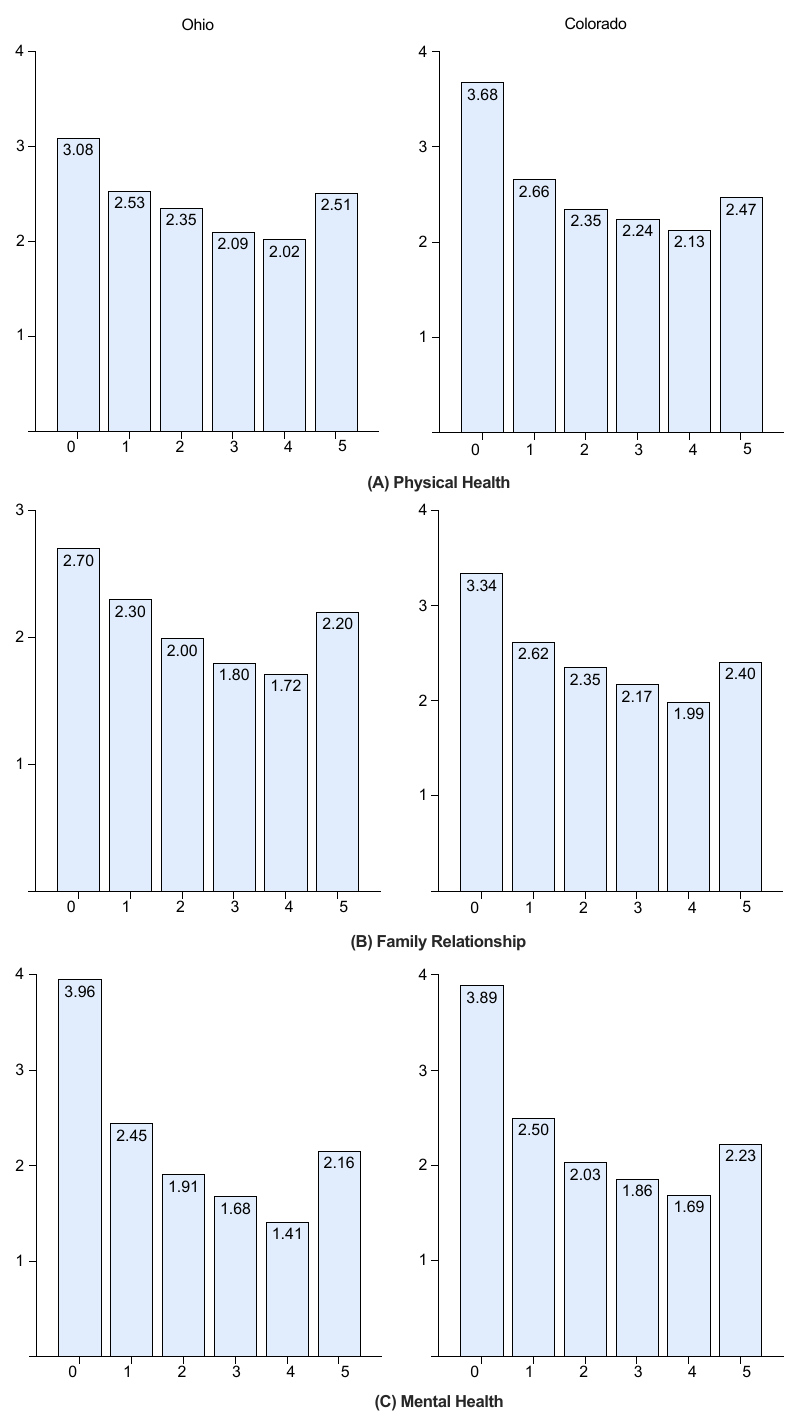}
    \caption{Prediction error count distributions (log scale) of Ohio and Colorado. (A) Physical Health, (B) Family Relationship, (C) Mental Health Crisis. Data instances with a prediction error count equal to 5 will be identified as potential mistakes.}
    \label{fig:figure2}
\end{figure}

Figure~\ref{fig:figure2} shows the prediction error count distributions (log scale) for two illustrative states, Ohio and Colorado. We identified problematic instances by establishing a threshold on the prediction error counts, setting it at a value of 5. Table~\ref{tab:statistics} offers a detailed statistical summary of these problematic data instances. For Ohio, our problematic instance discovery identified 159 potential mistakes out of 1,077 Family Relationship Crisis annotations (14.8\%), 324 out of 2,328 Physical Health Crisis annotations (13.9\%), and 143 out of 9,654 Mental Health Crisis annotations (1.5\%). For Colorado, our method detected 254 potential mistakes out of 3,315 Family Relationship Crisis annotations (7.7\%), 294 out of 6,019 Physical Health Crisis annotations (4.9\%), and 168 out of 8,534 Mental Health Crisis annotations (2.0\%).

\begin{table}
    \vspace{1em}
    \centering
    \caption{Statistics of the identified problematic data instances. PMs - Potential Mistakes.}
    \label{tab:statistics}
    \begin{tabular}{lrrrrrrrrr}
    \toprule
    State & \multicolumn{3}{c}{Physical Health} & \multicolumn{3}{c}{Family Relationship} & \multicolumn{3}{c}{Mental Health} \\
    \cmidrule(rl){2-4}\cmidrule(rl){5-7}\cmidrule(rl){8-10}
    & Total & PMs & (\%) & Total & PMs & (\%) & Total & PMs & (\%)\\
    \midrule
    Ohio & 1,077 & 159 & 14.8 & 2,328 & 324 & 13.9 & 9,654 & 143 & 1.5 \\
    Colorado & 3,315 & 254 & 7.7 & 6,019 & 294 & 4.9 & 8,534 & 168 & 2.0 \\
    \bottomrule
    \end{tabular}
    \vspace{1em}
\end{table}

\subsection{Verifying Annotation Consistency}\label{verifying-annotation-consistency}

\begin{figure}[hbt!]
    \centering
    \includegraphics[width=.6\textwidth]{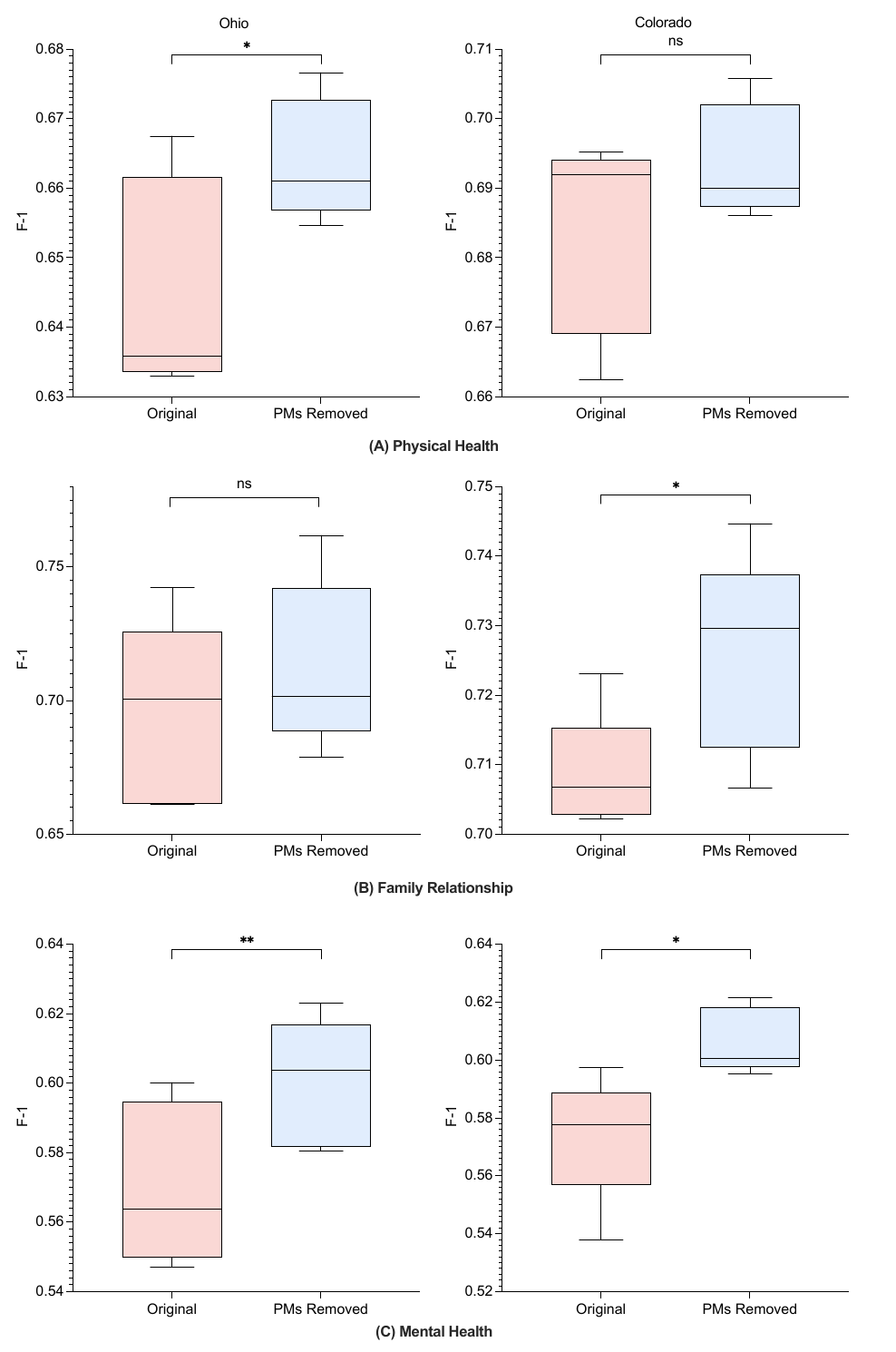}
    \caption{Comparison of \fscore scores between models trained using `Original' (before removing the identified potential mistakes), and `PMs Removed' (after removing the identified potential mistakes). (A) Physical Health, (B) Family Relationship, (C) Mental Health Crisis. PMs - Potential Mistakes. The asterisk indicates statistical significance.}
    \label{fig:figure3}
\end{figure}

Figure~\ref{fig:figure3} visually represents our annotation consistency verification results. We investigated whether these potential mistakes adversely affect the annotation consistency between the target and other states. To assess this impact, we re-trained our classifiers, excluding the problematic instances we identified, and compared the performances on other states' test set. To provide a reference point, we introduced a random baseline, which randomly removed the same number of instances from the training set as the problematic ones. We conducted these experiments with different random seeds five times, subjecting the results to T-tests for statistical analysis. Detailed \fscore scores are available in Table~\ref{tab:table2}.

After removing potential mistakes from Ohio's data, we observed notable improvements in the average micro \fscore scores for each crisis on other states' test sets. In contrast, the random baseline resulted in smaller performance gains for all three crises. Specifically, for Family Relationship Crisis, the score increased from 0.695 to 0.713 after removing the potential mistakes, compared to an increase from 0.695 to 0.701 with random dropping. For Physical Health Crisis, it improved from 0.645 to 0.664 after removing the potential mistakes, as opposed to an increase from 0.645 to 0.654 with random dropping. For Mental Health Crisis, it rose from 0.571 to 0.600 when the potential mistakes were removed, in contrast to an increase from 0.571 to 0.585 with random dropping.

Similar trends were observed in Colorado. After removing potential mistakes, the average micro \fscore score for Family Relationship Crisis on the test set of other states increases from 0.705 to 0.726, compared to an increase from 0.705 to 0.714 with random dropping. For Physical Health Crisis, it increased from 0.684 to 0.694, in contrast to an increase from 0.684 to 0.690 with random dropping. For Mental Health Crisis, it rose from 0.574 to 0.607, compared to an increase from 0.574 to 0.587 with random dropping.

These consistent trends observed across two states and all three crises suggest that removing potential mistakes helps align the label annotations of the target state with those of other states, underscoring the effectiveness of our approach in identifying annotation mistakes.

\begin{table}[hbt!]
    \centering
    \caption{Comparison of \fscore scores between models trained using `Original' (before removing the identified potential mistakes), `PMs Randomly Dropped' (after randomly dropping the identified potential mistakes), and `PMs Removed' (after removing the identified potential mistakes). PMs - Potential Mistakes.}
    \label{tab:table2}
    \small
    \begin{tabular}{l*{10}{c}}
    \toprule
     &  \multicolumn{5}{c}{Ohio} & \multicolumn{5}{c}{Colorado} \\
    \cmidrule(rl){2-6} \cmidrule(rl){7-11}
    Crisis &  1 &  2 &  3 &  4 &  5 &  1 &  2 &  3 &  4 &  5 \\
    \midrule
    Family Relationship\\
    \hspace{1em} Original & 0.656 & 0.636 & 0.633 & 0.667 & 0.634 & 0.707 & 0.723 & 0.703 & 0.702 & 0.707 \\
    \hspace{1em} PMs Randomly Dropped & 0.632 & 0.649 & 0.657 & 0.665 & 0.665 & 0.718 & 0.719 & 0.717 & 0.709 & 0.707 \\
    \hspace{1em} PMs Removed & 0.659 & 0.661 & 0.669 & 0.677 & 0.655 & 0.729 & 0.745 & 0.718 & 0.730 & 0.707 \\
    Mental Health\\ 
    \hspace{1em} Original & 0.553 & 0.547 & 0.600 & 0.564 & 0.589 & 0.597 & 0.576 & 0.578 & 0.538 & 0.580 \\
    \hspace{1em} PMs Randomly Dropped & 0.563 & 0.590 & 0.598 & 0.580 & 0.593 & 0.556 & 0.602 & 0.581 & 0.590 & 0.604 \\
    \hspace{1em} PMs Removed & 0.623 & 0.604 & 0.580 & 0.611 & 0.583 & 0.601 & 0.615 & 0.600 & 0.621 & 0.595\\
    Physical Health\\
    \hspace{1em} Original & 0.661 & 0.661 & 0.700 & 0.742 & 0.709 & 0.692 & 0.676 & 0.693 & 0.695 & 0.662 \\
    \hspace{1em} PMs Randomly Dropped & 0.697 & 0.677 & 0.742 & 0.697 & 0.692 & 0.689 & 0.680 & 0.687 & 0.699 & 0.695 \\
    \hspace{1em} PMs Removed & 0.723 & 0.698 & 0.762 & 0.679 & 0.701 & 0.706 & 0.690 & 0.686 & 0.688 & 0.698 \\
    \bottomrule
    \end{tabular}
    \vspace{1em}
\end{table}

\subsection{Rectifying Problematic Data }\label{rectifying-problematic-data}

\begin{figure}[hbt!]
    \centering
    \includegraphics[width=\textwidth]{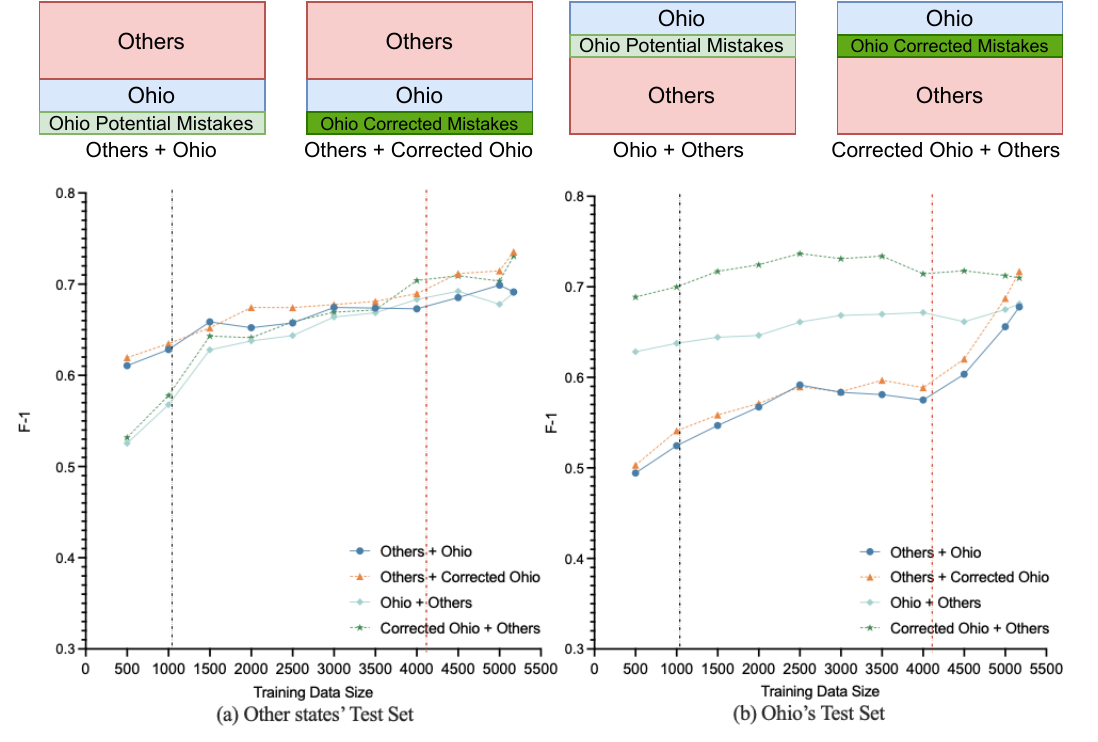}
    \caption{Comparisons of average micro \fscore scores for Family Relationship Crisis when we gradually feed more training data in an incremental manner to the model, (a) Other states’ test set, (b) Ohio's test set. In either subplot, the black vertical dashed line on the left denotes when Ohio's data have all been fed to the model for training data \textbf{Ohio+Others} and \textbf{CorrectedOhio+Others}, while the red vertical dashed line on the right denotes when we start to feed Ohio’s data to the model for \textbf{Others+Ohio} and \textbf{Others+CorrectedOhio}.}
    \label{fig:figure4}
\end{figure}

We recruited two annotators to manually identify and correct the actual mis-labelings in the 159 potential mistakes in Ohio's Family Relationship Crisis annotations. The actual mis-labelings are defined as instances where the two annotators identify ground truth annotations as incorrect. Two annotators received training on annotating labels following the NVDRS coding manual and resolved disagreements through discussion, achieving a high Inter-Annotator Agreement (IAA) of 0.893 (Kappa value). Among the 159 potential mistakes, 89 were confirmed as actual mis-labelings. These included 87 instances where the Family Relationship Crisis labels were incorrectly labeled as `0' in the ground truth annotations, and 2 instances where the labels were mistakenly labeled as `1' in the ground truth annotations.

Figure~\ref{fig:figure4} shows the changes in the average micro \fscore scores during the incremental training process. In Figure~\ref{fig:figure4}(a), the model performance on the test set of other states exhibits significant improvement when the corrected data is fed to the model at the beginning of the training process (to the left of the black vertical dashed line), and at the end of the training process (to the right of the red vertical dashed line). The label correction boosts the eventual average micro \fscore score on the test set of other states from 0.691 to 0.733. A substantial improvement can also be observed on Ohio's test set in Figure~\ref{fig:figure4}(b). The label correction enhances the eventual average micro \fscore score on Ohio's test set from 0.679 to 0.714. This result demonstrates that, after correction, the corrected data instances benefit the model performances on both the test sets of other states and the target state, regardless of whether they are fed to the model at the beginning or the end of the training process. This highlights the importance of accurate and consistent annotations in improving the performance of classifiers across different states, ultimately leading to more reliable results and better model outcomes.

\subsection{Risk of Bias Analysis}\label{risk-of-bias-analysis}

\begin{table}[hbt!]
    \centering
    \footnotesize
    \noindent
    \caption{Odds Ratio for crisis between youth vs. adults, Blacks vs. whites, and females vs. males. OR - Odds Ratio. CI - Confidence Interval.}
    \begin{tabular}{l*{3}{r@{\hspace{9pt}}r@{\hspace{12pt}}r}}
    \toprule
    & \multicolumn{9}{c}{Ohio} \\
    \cmidrule(rl){2-4} \cmidrule(rl){5-7} \cmidrule(rl){8-10} 
    Crisis & Youth & Adults & OR[95\% CI] & Black & White & OR[95\% CI] & Female & Male & OR[95\% CI] \\
    \midrule
    Family Relation.\\
    \hspace{.5em} Original & 250	& 827 & 1.01[0.75;1.36] & 103	& 947 & 1.53[1.01;2.32] & 310 & 767 & 1.12[0.85;1.48] \\
    \hspace{.5em} Random Drop & 216 & 702 & 1.04[0.75;1.44] & 84 & 810 & 1.56[0.99;2.47] & 268 & 650 & 1.16[0.86;1.57] \\
    \hspace{.5em} Our Method & 216 & 702 & 1.05[0.76;1.45]	& 90 & 805 & 1.51[0.97;2.36] & 256 & 662 & 1.06[0.78;1.44] \\
    Mental Health\\
    \hspace{.5em} Original & 1,218 & 8,436 & 0.79[0.43;1.44] & 859 & 8,638 & 0.99[0.51;1.91] & 2,814 & 6,840 & 0.95[0.63;1.43] \\
    \hspace{.5em} Random Drop & 1,201 & 8,310 & 0.79[0.43;1.46] & 847 & 8,510 & 1.00[0.52;1.92] & 2,767 & 6,744 & 0.96[0.64;1.45] \\
    \hspace{.5em} Our Method & 1,204 & 8,307 & 0.81[0.44;1.48] & 846 & 8,510 &1.00[0.52;1.94] & 2,780 & 6,731 & 0.98[0.65;1.47] \\
    Physical Health\\
    \hspace{.5em} Original & 30 & 2,298 & 0.52[0.18;1.51] & 84 & 2,221	& 1.01[0.60;1.70] & 442 & 1,886 & 0.63[0.48;0.82] \\
    \hspace{.5em} Random Drop & 26 & 1,978 & 0.64[0.22;1.88] & 69 & 1,914 & 0.64[0.33;1.24] & 369 & 1,635 & 0.60[0.44;0.80] \\
    \hspace{.5em} Our Method & 28 & 1,976 & 0.34[0.08;1.45] & 73 & 1,910 & 0.79[0.41;1.52] & 387 & 1,617 & 0.49[0.35;0.68] \\
    \midrule
    & \multicolumn{9}{c}{Colorado} \\
    \cmidrule(rl){2-4} \cmidrule(rl){5-7} \cmidrule(rl){8-10} 
    Crisis & Youth & Adults & OR[95\% CI] & Black & White & OR[95\% CI] & Female & Male & OR[95\% CI] \\
    \midrule
    Family Relation.\\
    \hspace{.5em} Original & 684	& 2,631	& 1.38[1.07;1.80] & 112 & 3,063 & 1.15[0.64;2.08] & 946 & 2,369 & 0.87[0.67;1.13] \\
    \hspace{.5em} Random Drop & 626 & 2,435 & 1.34[1.02;1.76] & 100 & 2,831 & 1.08[0.57;2.05] & 873 & 2,188 & 0.96[0.74;1.25] \\
    \hspace{.5em} Our Method & 613 & 2,448 & 1.43[1.01;2.01] & 102 & 2,827 & 1.35[0.64;2.83] & 879 & 2,182 & 0.88[0.63;1.24]\\
    Mental Health\\
    \hspace{.5em} Original & 1,237 & 7,297 & 0.89[0.59;1.33] & 199 & 8,034	& 0.68[0.49;0.93] & 2,743 & 5,791 & 0.20[0.03;1.42] \\
    \hspace{.5em} Random Drop & 1,205 & 7,161 & 0.88[0.58;1.34] & 195 & 7,874 & 0.67[0.48;0.93] & 2,688 & 5,678 & 0.20[0.03;1.45] \\
    \hspace{.5em} Our Method & 1,210 & 7,156 & 0.65[0.31;1.36]	& 197 & 7,874 & 0.51[0.07;3.70]	& 2,695	& 5,671	& 0.48[0.27;0.84] \\
    Physical Health\\
    \hspace{.5em} Original & 251 & 5,768	& 0.49[0.24;1.01]	& 120 & 5,736 & 0.50[0.38;0.66] & 1,720 & 4,299 & 0.12[0.02;0.86] \\
    \hspace{.5em} Random Drop & 235 & 5,490 & 0.67[0.39;1.14] & 115 & 5,454 & 0.65[0.53;0.81] & 1,640 & 4,085 & 0.08[0.01;0.60] \\
    \hspace{.5em} Our Method & 241 & 5,484 & 0.64[0.52;0.79] & 120 & 5,454	& 0.67[0.40;1.12]	& 1,641	& 4,084	& 0.08[0.01;0.58]\\
    \midrule
    \end{tabular}
    \label{tab:table3}
\end{table}

Table~\ref{tab:table3} shows the Odds Ratio (OR) comparisons between youth vs adults, Blacks vs whites, and females vs males. Notably, in relation to the Mental Health Crisis in Colorado, the OR for youth in the original NVDRS annotations (OR=0.89, 95\%CI=0.59-1.33) is similar to that in the annotations after random dropping (OR=0.88, 95\% CI=0.58-1.34). However, it differs from the OR in the annotations after removing the mistakes identified by our method (OR=0.65, 95\% CI=0.31-1.36). Similarly, the OR for the Black individuals in the original NVDRS annotations (OR=0.68, 95\% CI=0.49-0.93) is similar to that in the annotations after random dropping (OR=0.67, 95\% CI=0.48-0.93), but deviates from the OR in the annotations after removing the mistakes identified by our method (OR=0.51, 95\% CI=0.07-3.7). These observed OR differences suggest the necessity of our proposed mistake identification method.

\subsection{Discussion}\label{discussion}

This study touches on a previously unexplored area of uncovering annotation inconsistencies in unstructured death investigation notes and the resolution of misattributed suicide causes. To bridge this gap, we proposed an empirical NLP approach.

First, we demonstrated the existence of data annotation inconsistencies across states and showcased how these inconsistencies led to performance disparities in the developed crisis prediction systems. Our findings showed that after incorporating the data from the target state into the training set, prediction performance improvements were observed on the test sets of 49.8\% of the target states. Meanwhile, after adding the data from the target state to the training set, an average of 51.1\% of the target states showed prediction performance decreases on the test set of other states. The observed disparities in prediction performance using different combinations of training data uncovered the presence of annotation inconsistencies across different states.

Subsequently, we introduced a method to identify problematic instances responsible for these inconsistencies through a cross-validation-like paradigm. After thresholding the prediction error counts at a value of 5, our method found that for Ohio, 14.8\% of the annotations for Family Relationship Crisis, 13.9\% of the annotations for Physical Health Crisis, and 1.5\% of the annotations for Mental Health Crisis were potential mistakes; for Colorado, 7.7\% of the annotations for Family Relationship Crisis, 4.9\% of the annotations for Physical Health Crisis, and 2.0\% of the annotations for Mental Health Crisis were potential mistakes. We also analyzed several examples of errors to examine the nature of these discrepancies within the annotations.

To validate the effectiveness of our approach, we first illustrated that removing these problematic instances from the training set improved model performance and generalizability. Specifically, removing the problematic instances brought an average increase of 2.2\% to the average micro \fscore scores on the test set of other states, while the random baseline yielded a marginal 0.95\% improvement. Moreover, we manually rectified 159 discovered potential mistakes in Ohio's Family Relationship Crisis annotations, among which we found a total of 89 to be actual mis-labelings. After rectification, the average micro \fscore score on the test set of other states increased by 4.2\%, and the average micro \fscore score on Ohio's test set increased by 3.5\%.

Last but not least, to better understand the risk of bias in the data annotation, we employed logistic regression models to examine whether the relationship between the suicide-related circumstances and demographic variables (race, age, and sex) has changed as we removed the identified mistakes. Odds Ratios (ORs) were computed for various demographic subgroups (age, sex, race). We observed the OR differences between the original NVDRS annotations and annotations after removing the mistakes identified by our method. For example, in terms of the Mental Health Crisis in Colorado, the OR for youth in the original NVDRS annotations (OR=0.89, 95\%CI=0.59-1.33) is similar to that in the annotations after random dropping (OR=0.88, 95\% CI=0.58-1.34). Still, it differs from the OR in the annotations after removing the mistakes identified by our method (OR=0.65, 95\% CI=0.31-1.36).

While our study offers promising insights, it does have certain limitations. First, our problematic instance discovery method uses a cross-validation-like method, which can become computationally demanding as the dataset size increases. Secondly, although our proposed framework can work with various models, we only demonstrated the results utilizing BERT-based models. Several NLP tasks have recently showcased the effectiveness of Large Language Models (LLMs), which could be a potential area of exploration for future studies. Moreover, the choice of parameters, such as the number of folds and the threshold for error identification, can be further tuned using Grid Search for better results. Furthermore, our risk of bias analysis suggests that our proposed method will likely discover potential bias in the data annotations. Future research could help identify and address the possible bias. Lastly, although we demonstrated the effectiveness of manual label correction, automatic methods should be explored for scalability. Such an approach would provide a more practical means of improving annotation consistency across large datasets and various sources.

\section{Methods}\label{methods}
\subsection{Data Source}\label{data-source}

This work utilizes data from the National Violent Death Reporting System (NVDRS) dataset, covering 267,804 recorded suicide death incidents from 2003 to 2020 across all 50 U.S. states, Puerto Rico, and the District of Columbia \footnote{\url{https://www.cdc.gov/violenceprevention/datasources/nvdrs/index.html}}. Our research is approved by the NVDRS Restricted Access Database proposal.

Each incident instance is accompanied by two death investigation notes, one from the Coroner or Medical Examiner (CME) perspective and the other from the Law Enforcement (LE) perspective. The NVDRS contains over 600 unique data elements for each incident, including the identification of suicide crises - precipitating events contributing to the occurrence of suicides, that occurred within two weeks before a suicide death~\cite{Liu2023-ry}. Examples of suicide crises include Family Relationship, Physical Health, and Mental Health crises. These crises are manually extracted from both CME and LE reports by trained abstractors.

In this study, we present our methods and conduct experiments using three crises as illustrative examples: Family Relationship Crisis, Mental Health Crisis, and Physical Health Crisis (the state-wise statistics are detailed in Table~\ref{tab:table4}). These variables were selected for their higher prevalence of positive instances in the NVDRS dataset and their poor classification scores, as demonstrated in prior work~\cite{Wang2023-ov}. Definitions and examples of these three crises can be found in Supplementary Table~\ref{tab:supplement_table1}. We also addressed the positive/negative class imbalance in the NVDRS dataset through data pre-processing. First, states with fewer than 10 positive instances were excluded to ensure adequate training data. Next, for each crisis, we created a balanced class distribution for every state by keeping the positive instances intact and down-sampling the negative instances, ensuring an equal number of both.

\begin{table}[]
    \centering
    \footnotesize
    \vspace{-1.5em}
    \caption{State-wise data statistics.}
    \label{tab:table4}
    \begin{tabular}{l*{10}{r}}
    \toprule
    State & \multicolumn{3}{c}{Family Relationship} & \multicolumn{3}{c}{Mental health} & \multicolumn{3}{c}{Physical Health} \\
    \cmidrule(rl){2-4} \cmidrule(rl){5-7} \cmidrule(rl){8-10} 
    & Positive & Negative & Total & Positive & Negative & Total & Positive & Negative & Total \\
    \midrule
    Alabama & 4 & 327 & 331 & 1 & 491 & 492 & 8 & 159 & 167 \\
    Alaska & 102 & 213 & 315 & 1 & 622 & 623 & 50 & 177 & 227 \\ 
    Arizona & 198 & 1,116 & 1,314 & 69 & 2,762 & 2,831 & 117 & 552 & 669 \\ 
    Arkansas & 1 & 50 & 51 & 1 & 107 & 108 & 0 & 34 & 34 \\
    California & 179 & 884 & 1,063 & 113 & 2,964 & 3,077 & 117 & 233 & 350 \\
    Colorado & 536 & 2,779 & 3,315 & 204 & 8,330 & 8,534 & 324 & 5,695 & 6,019 \\
    Connecticut & 1 & 627 & 628 & 1 & 1,177 & 1,178 & 1 & 152 & 153 \\
    Delaware & 15 & 88 & 103 & 7 & 222 & 229 & 7 & 36 & 43 \\
    District of Columbia & 10 & 13 & 23 & 9 & 160 & 169 & 14 & 17 & 31 \\
    Florida & 0 & 0 & 0 & 0 & 0 & 0 & 0 & 1 & 1 \\
    Georgia & 240 & 1,113 & 1,353 & 29 & 2,845 & 2,874 & 104 & 577 & 681 \\
    Hawaii & 3 & 174 & 177 & 3 & 303 & 306 & 2 & 85 & 87 \\ 
    Idaho & 50 & 36 & 86 & 43 & 126 & 169 & 25 & 31 & 56 \\
    Illinois & 161 & 718 & 879 & 39 & 2,506 & 2,545 & 94 & 313 & 407 \\
    Indiana & 255 & 495 & 750 & 68 & 1,599 & 1,667 & 103 & 205 & 308 \\
    Iowa & 233 & 324 & 557 & 301 & 1,283 & 1,584 & 75 & 302 & 377 \\
    Kansas & 67 & 761 & 828 & 19 & 1,396 & 1,415 & 87 & 233 & 320 \\
    Kentucky & 142 & 1,212 & 1,354 & 84 & 1,783 & 1,867 & 89 & 358 & 447 \\
    Louisiana & 86 & 288 & 374 & 14 & 894 & 908 & 54 & 250 & 304 \\
    Maine & 76 & 257 & 333 & 38 & 612 & 650 & 50 & 101 & 151 \\
    Maryland & 113 & 412 & 525 & 42 & 4,284 & 4,326 & 103 & 173 & 276 \\
    Massachusetts & 283 & 598 & 881 & 44 & 3,061 & 3,105 & 149 & 239 & 388 \\
    Michigan & 216 & 2,849 & 3,065 & 30 & 5,279 & 5,309 & 83 & 1,268 & 1,351 \\
    Minnesota & 184 & 669 & 853 & 213 & 2,618 & 2,831 & 133 & 544 & 677 \\
    Mississippi & 4 & 24 & 28 & 0 & 19 & 19 & 4 & 14 & 18 \\
    Missouri & 121 & 487 & 608 & 41 & 1,470 & 1,511 & 83 & 171 & 254 \\
    Montana & 2 & 154 & 156 & 3 & 186 & 189 & 2 & 53 & 55 \\
    Nebraska & 39 & 137 & 176 & 59 & 245 & 304 & 20 & 66 & 86 \\
    Nevada & 68 & 416 & 484 & 18 & 734 & 752 & 54 & 171 & 225 \\
    New Hampshire & 38 & 270 & 308 & 15 & 993 & 1,008 & 2 & 151 & 153 \\
    New Jersey & 217 & 623 & 840 & 114 & 2,320 & 2,434 & 189 & 193 & 382 \\
    New Mexico & 141 & 622 & 763 & 50 & 1,041 & 1,091 & 67 & 439 & 506 \\
    New York & 83 & 815 & 898 & 60 & 3,887 & 3,947 & 54 & 380 & 434 \\
    North Carolina & 640 & 1,929 & 2,569 & 104 & 5,527 & 5,631 & 675 & 280 & 955 \\
    North Dakota & 17 & 42 & 59 & 10 & 152 & 162 & 14 & 46 & 60 \\
    Ohio & 470 & 607 & 1,077 & 95 & 8,439 & 8,534 & 300 & 2,028 & 2,328 \\
    Oklahoma & 299 & 825 & 1,124 & 59 & 2,784 & 2,843 & 104 & 793 & 897 \\
    Oregon & 161 & 1,108 & 1,269 & 42 & 2,729 & 2,771 & 76 & 402 & 478 \\
    Pennsylvania & 288 & 930 & 1,218 & 13 & 2,749 & 2,762 & 65 & 346 & 411 \\
    Rhode Island & 29 & 112 & 141 & 20 & 523 & 543 & 22 & 75 & 97 \\
    South Carolina & 105 & 1,224 & 1,329 & 72 & 1,964 & 2,036 & 71 & 364 & 435 \\
    South Dakota & 0 & 16 & 16 & 0 & 56 & 56 & 0 & 12 & 12 \\
    Tennessee & 95 & 129 & 224 & 11 & 652 & 663 & 78 & 54 & 132 \\
    Texas & 29 & 103 & 132 & 46 & 365 & 411 & 48 & 51 & 99 \\
    Utah & 676 & 760 & 1,436 & 191 & 2,641 & 2,832 & 550 & 348 & 898 \\
    Vermont & 42 & 101 & 143 & 36 & 421 & 457 & 25 & 34 & 59 \\
    Virginia & 493 & 1,315 & 1,808 & 352 & 4,817 & 5,169 & 621 & 444 & 1,065 \\
    Washington & 570 & 697 & 1,267 & 166 & 2,762 & 2,928 & 348 & 560 & 908 \\
    West Virginia & 12 & 204 & 216 & 1 & 350 & 351 & 1 & 100 & 101 \\
    Wisconsin & 297 & 978 & 1,275 & 17 & 2,811 & 2,828 & 169 & 537 & 706 \\
    Wyoming & 0 & 51 & 51 & 0 & 85 & 85 & 1 & 13 & 14 \\
    Puerto Rico & 20 & 156 & 176 & 19 & 600 & 619 & 15 & 64 & 79 \\
    \bottomrule
    \end{tabular}
\end{table}

\subsection{Validate Annotation Inconsistency}\label{validate-annotation-inconsistency}

\begin{figure}[hbt!]
    \centering
    \includegraphics[width=\textwidth]{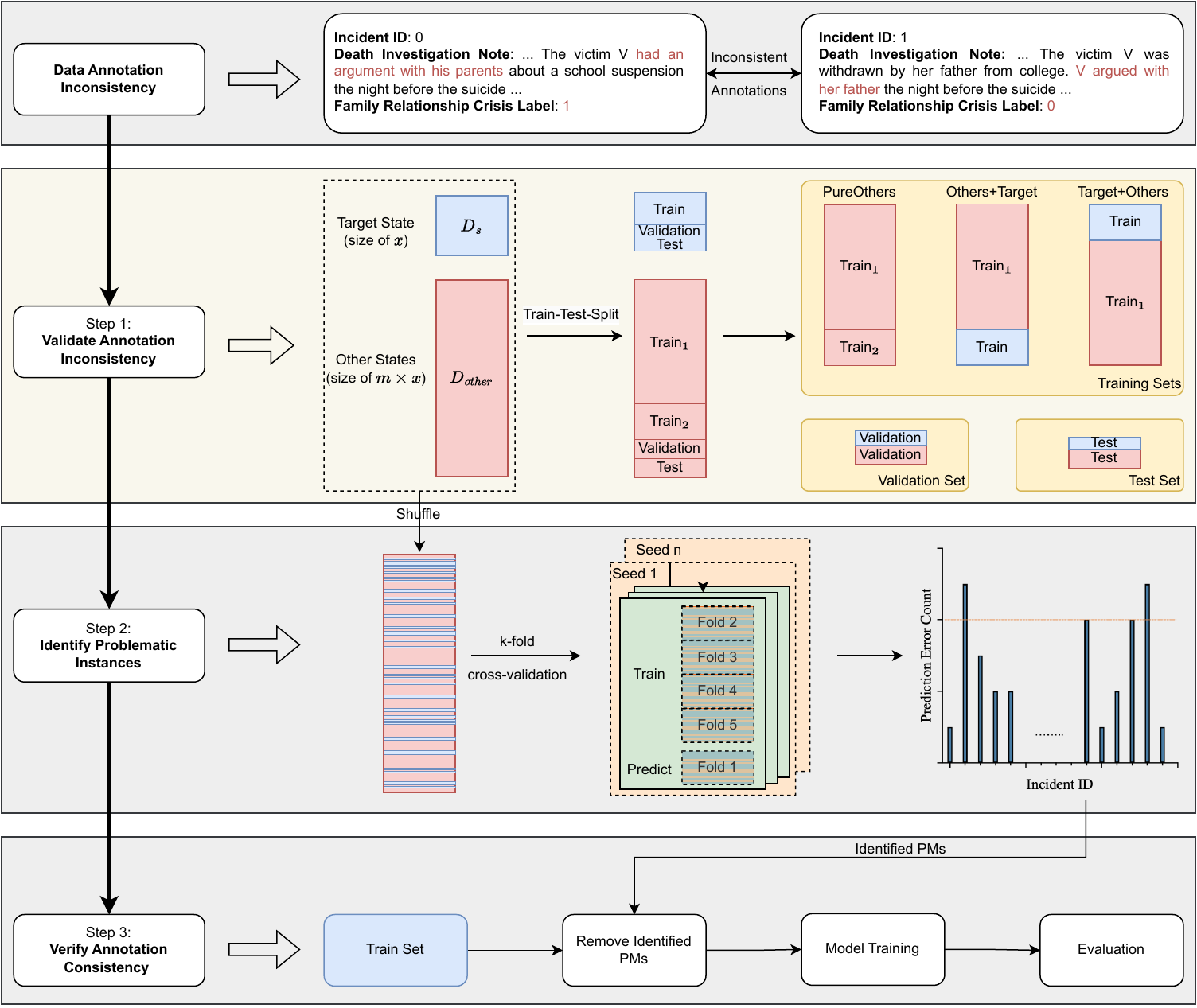}
    \caption{Annotation inconsistency example and our proposed framework. In Step 1, the size of other states' $Train_2$ set equals the size of the target state's $Train$ set, ensuring the three new training sets are of the same size. In Step 2, the \(k\)-fold cross-validation procedure is repeated n times using different random seeds. For each data instance, we recorded its prediction error counts, and eventually identified the problematic instances by thresholding the prediction error counts. PMs - Potential Mistakes.}
    \label{fig:figure5}
\end{figure}

We assume that if the label annotations for two data subsets are consistent, they should be equivalently predictive of each other. In practical terms, given a dataset \(D\), if we train a model using one of its subsets to predict the remaining portion, we anticipate observing a comparable evaluation performance for both subsets.

Based on this assumption, we first explore whether the label annotations in the target state \(s\) are consistent with those in all other states (Step 1 in Figure~\ref{fig:figure5}). Specifically, given the annotated data of target state \(D_{s} \subset D\) (where \(D_{s}\) has a size of \(x\)), we sample \(m\) exclusive subsets (each with a size of \(x\)) from the annotated data of other states, denoted as \(D_{other}\). It is worth noting that \(D_{s} \cap D_{other} = \phi\) .

We then split \(D_{s}\) and \(D_{other}\) into training, validation, and test sets, respectively, with a ratio of 8:1:1, and construct three different training sets of the same size: (1) \textbf{PureOthers} exclusively comprising samples from states other than the target state, (2) \textbf{Others+Target} combining samples of other states with samples of the target state, and (3) \textbf{Target+Others} similarly combining samples of the target and samples of other states in order. Our goal is to compare the classification performances between different training set combinations. Specifically, we assess the inconsistencies between every state and other states in the annotations of Physical Health, Family Relationship, and Mental Health crises. To quantify the inconsistency, we compute the $\Delta \fscore$'s on the test sets for both the target state and other states. The inconsistency is measured as the difference between the average \fscore score of models trained using mixed training data (\textbf{Others+Target} and \textbf{Target+Others}) and the \fscore score of the model trained solely on data from other states (\textbf{PureOthers}):
\begin{align} 
\Delta \fscore &=  \text{Difference}(\fscore_\text{Mixed} - \fscore_{\text{PureOthers}}) \\ 
\fscore_\text{Mixed} &=  \text{Mean}(\fscore_\text{Others + Target},\fscore_\text{Target + Others})
\end{align}

When incorporating the data from the target state into training, a larger positive $\Delta \fscore$ on the test set of the target state, accompanied by a smaller negative $\Delta \fscore$ on the test set of other states, indicates a more pronounced annotation inconsistency between the target state and other states.

\subsection{Identify Problematic Instances}\label{identify-problematic-instances}

To identify problematic data instances in the target state which may have contributed to the label inconsistencies between \(D_{s}\) and \(D_{other}\), we introduce an approach inspired by \(k\)-fold cross-validation (Step 2 in Figure~\ref{fig:figure5}). Our method involves the following steps: we concatenate \(D_{s}\) and \(D_{other}\) into one set, we randomly shuffle the data to ensure it is well-mixed, and we divide the shuffled dataset into \(k\) folds. Each unique fold is treated as a hold-out set, while the remaining \(k - 1\) folds serve as the training set.

Throughout this process, each individual data sample gets assigned to a specific fold where it remains for the duration of the cross-validation. This ensures that each data sample is utilized once in the hold-out set and contributes to training the model \(k - 1\) times. For each data sample in the hold-out set, we compare the model's prediction to the ground truth label and count the number of discrepancies.

To reduce randomness, we iteratively repeat the \(k\)-fold cross-validation procedure multiple times, employing different random seeds on each iteration. Then, we apply a thresholding mechanism to the count of prediction errors for each data sample in \(D_{s}\). This thresholding enables us to effectively identify and flag problematic data instances.

\subsection{Verify Annotation Consistency}\label{verify-annotation-consistency}

\begin{figure}
    \centering
    \vspace{1em}
    \includegraphics[width=\textwidth]{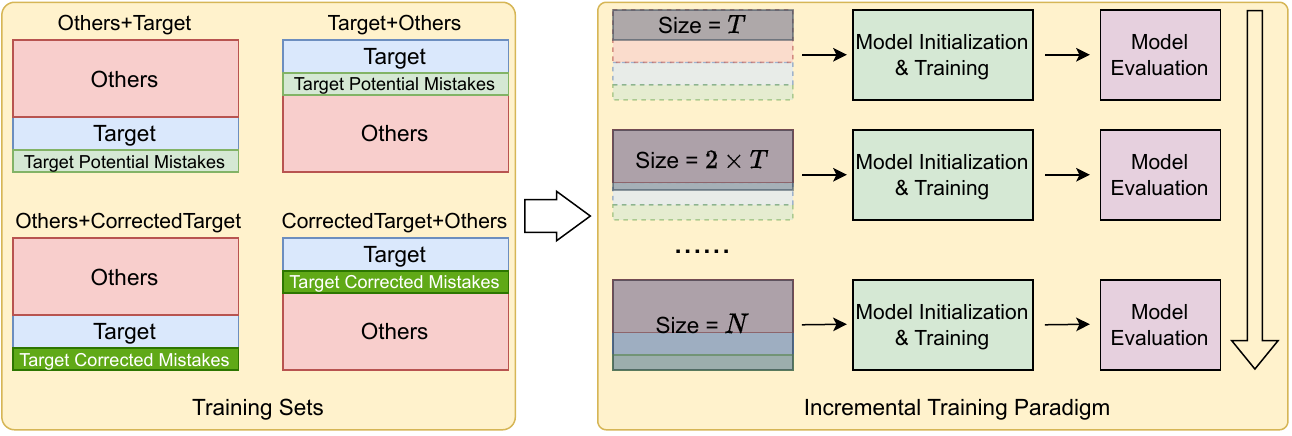}
    \caption{Illustration of the proposed incremental training paradigm. $T$ is the step size, and $N$ is the total number of data instances in the target state $s$ and other states. Four combinations of training data are shown on the left. For each combination of training data, we feed it to the incremental training paradigm, as shown on the right.}
    \label{fig:figure6}
    \vspace{1em}
\end{figure}

Once we've identified the problematic data instances in \(D_{s}\), our next step is to evaluate whether these potential mistakes negatively impact the model's performance. To this end, we remove potential mistakes from the training dataset. By systematically removing these instances and re-training the model, we can measure the impact of the potential mistakes on the model's performance (Step 3 in Figure~\ref{fig:figure5}). On a separate front, our efforts extend to manual correction of potential mistakes. Our objective is to show how consistent annotations can enhance the performance of classifiers. We employ an incremental training paradigm to demonstrate this. After identifying the potential mistakes, two annotators manually revisited the problematic data instances, and corrected the labels if necessary. We aim to validate the label consistency of the corrected data using an incremental training paradigm with four training sets (Figure~\ref{fig:figure6}): \textbf{Others+Target}, comprising the data from other states and the original data from target state; \textbf{Others+CorrectedTarget}, comprising the data from other states and the data from target state after correction; \textbf{Target+Others}, comprising the original data from the target state and the data from other states, and \textbf{CorrectedTarget+Others} with the data from the target state after correction and the data from other states.

For each training set, we progressively incorporate more training samples in an incremental manner using a step size of \(T\), to have a finer-grained view of how the corrected data impact the model performance. We train the classification models and analyze the performances on the test set. This process helps validate the label consistency and the effectiveness of the corrected data. We repeat all experiments \(n\)=5 times using different random seeds.

\subsection{Risk of Bias Analysis}\label{risk-of-bias-analysis-1}

To better understand the risk of bias in the data annotation, we employed logistic regression models to examine whether the relationship between the suicide-related circumstances and demographic variables (i.e., race, age, and sex) has changed as we removed the identified mistakes. One distinct logistic regression model was developed for each suicide-related circumstance.

Specifically, the predictor variable represented the specific comparison group (i.e., Black, youth (age under 24), female) and was coded as 1. This was then contrasted with the reference group (i.e., white, adult, male), coded as 0. We calculated the ORs for each comparison group using the coefficient estimate affiliated with the predictor variable obtained from the corresponding logistic regression model. The OR quantifies the likelihood of the specific circumstance occurring in a comparison group versus the reference group. The OR is computed as follows:

\begin{equation}
    OR = e^{\text{Coefficient Estimate for the Comparison Group}}
\end{equation}

ORs greater than 1 indicate that the comparison group had higher circumstance rates than the reference group. We further calculated a 95\% confidence interval (CI) for each OR based on the standard error of the coefficient estimate and the Z-score as follows:

\begin{align}
    \text{Lower CI Bound} &= e^{\text{Coefficient Estimate}} - Z \times \text{Standard Error}\\
    \text{Higher CI Bound} &= e^{\text{Coefficient Estimate}} + Z \times \text{Standard Error}
\end{align}

For two illustrative states (Ohio and Colorado), we computed the ORs of each circumstance variable in three sets of annotations: the original annotations from the NVDRS, the annotations after removing the mistakes identified by our method, and the annotations after randomly dropping the same number of instances as the identified mistakes. By comparing the ORs for the same subgroup in different sets of annotations, we can examine whether the relationship between the suicide-related circumstances and demographic variables has changed.

\subsection{Experiment Settings}\label{experiment-settings}

In this study, we used BioBERT as our backbone model~\cite{Lee2020-id}, known for its state-of-the-art performance, as demonstrated in our prior study~\cite{Wang2023-ov}. BioBERT works with sequences of up to 512 tokens, producing 768-dimensional representations. About 5.1\% of the NVDRS data have input length longer than 512 tokens, and they were truncated before being fed to BioBERT. We framed suicide crisis detection as a text classification problem by feeding concatenated CME and LE notes into BioBERT and training it to classify whether a suicide crisis of interest is mentioned in the text. We appended a fully connected layer on top of BioBERT for classification.

For experiments, we sampled \(m = 4\) exclusive subsets from the annotated data of other states. We conducted the experiments \(n = 5\) times, each with different random seeds, and reported the range of micro \fscore scores with the average. For problematic instance discovery, we chose \(k = 5\) for \(k\)-fold cross-validation. A higher frequency of discrepancies between prediction results and ground truth labels increases the probability of an incorrect ground truth label. We set the threshold at 5, effectively minimizing the number of false potential mistakes. We selected Physical Health, Family Relationship, and Mental Health as illustrative crises, Ohio and Colorado as illustrative states in this study. Additional details on crisis and state selections, as well as training, are provided in Supplementary~\ref{supplementaryA}.

\section{Conclusions}\label{conclusions}

The presence of data annotation inconsistencies in NVDRS's death investigation notes not only hampers our understanding of suicide causes but also impedes the development, implementation, and evaluation of effective strategies, programs, and policies aimed at preventing suicide. In this work, we proposed an empirical NLP approach to detect the data annotation inconsistencies in the NVDRS, identify problematic instances causing these inconsistencies, and further verify the effectiveness of correcting problematic instances. Experiment results showcase the capabilities and generalizability of our approach and suggest the limitations of this work. We intend to refine and expand our methodology to address data annotation inconsistencies across diverse data sources. Additionally, we advocate for establishing more stringent annotation guidelines and quality control measures to ensure the consistent and reliable annotation of datasets. By enhancing the accuracy and consistency of annotations within these datasets, we can elevate the performance and reliability of NLP models. This, in turn, equips scientists and policymakers with the means to improve the annotation accuracy for the NVDRS data, and then fundamentally supports discovering the true suicide causes, and eventually contributes to suicide prevention.

\section*{Code Availability}\label{code-availability}
We have made our code publicly available at \url{https://github.com/bionlplab/2024_npjDM_Inconsistency_Detection}.

\section*{Data Availability}\label{data-availability}
The dataset analyzed during this study, NVDRS RAD, is available by request for users meeting certain eligibility criteria. This is because NVDRS contains confidential information that could lead to accidental disclosure of the identity of suspects and victims. CDC protects these data by requiring users to meet certain eligibility requirements and to take steps necessary to ensure the security of data, preserve confidentiality, and prevent unauthorized access. Researchers can apply to access NVDRS as instructed here: \url{https://www.cdc.gov/violenceprevention/datasources/nvdrs/dataaccess.html}.

\section*{Acknowledgement}\label{acknowledgement}

This work was supported by the NSF CAREER Award No. 2145640, Amazon Research Award, and National Institutes of Health Aim-Ahead Award No. OT2OD032581.

\section*{Author Contributions}\label{author-contributions}
S.W, Y.X, Y.P contributed to the conception of the study and study design; S.W, Y.X, Y.P contributed to acquisition of the data; S.W, Y.Z, Y.X, Y.P contributed to analysis and interpretation of the data; Z.H, C.T, Y.X, Y.D, J.G, Y.P provided strategic guidance; S.W, Y.P contributed to paper organization and team logistics; S.W, Y.Z, Z.H, C.T, Y.X, Y.D, J.G, Y.P contributed to drafting and revising the manuscript.

\section*{Competing Interests}\label{competing-interests}
All authors declare no financial or non-financial competing interests.

\bibliographystyle{medline}
\bibliography{ref}

\begin{thebibliography}{10}

\bibitem{Liu2023-ry}
Liu GS, Nguyen BL, Lyons BH, Sheats KJ, Wilson RF, Betz CJ, Fowler KA.
\newblock Surveillance for Violent Deaths - National Violent Death Reporting System, 48 States, the District of Columbia, and Puerto Rico, 2020.
\newblock MMWR Surveill Summ. 2023 May;72(5):1--38.

\bibitem{Hollenstein2016-vw}
Hollenstein N, Schneider N, Webber B.
\newblock Inconsistency Detection in Semantic Annotation.
\newblock In: Proceedings of the Tenth International Conference on Language Resources and Evaluation ({{LREC}'16}). Portoro{\v z}, Slovenia: European Language Resources Association (ELRA); 2016. p. 3986--3990.

\bibitem{Wang2023-ov}
Wang S, Dang Y, Sun Z, Ding Y, Pathak J, Tao C, Xiao Y, Peng Y.
\newblock An {NLP} approach to identify {SDoH-related} circumstance and suicide crisis from death investigation narratives.
\newblock J Am Med Inform Assoc. 2023 Apr;30(8):1408--1417.

\bibitem{Kveton2002-nf}
Kv{\u e}to{\v n} P, Oliva K.
\newblock ({Semi-)Automatic} Detection of Errors in {PoS-Tagged} Corpora.
\newblock In: {COLING} 2002: The 19th International Conference on Computational Linguistics; 2002. .

\bibitem{Ma2001-vx}
Ma Q, Lu BL, Murata M, Ichikawa M, Isahara H.
\newblock {On-Line} Error Detection of Annotated Corpus Using Modular Neural Networks.
\newblock In: Proceedings of the International Conference on Artificial Neural Networks. ICANN '01. Berlin, Heidelberg: Springer-Verlag; 2001. p. 1185--1192.

\bibitem{Ule2004-ri}
Ule T, Simov K.
\newblock Unexpected Productions May Well be Errors.
\newblock In: Proceedings of the Fourth International Conference on Language Resources and Evaluation ({{LREC}{'}04}). Lisbon, Portugal: European Language Resources Association (ELRA); 2004. p. 1795--1798.

\bibitem{Loftsson2009-dp}
Loftsson H.
\newblock Correcting a {PoS-tagged} corpus using three complementary methods.
\newblock In: Proceedings of the 12th Conference of the European Chapter of the Association for Computational Linguistics. EACL '09. USA: Association for Computational Linguistics; 2009. p. 523--531.

\bibitem{Kato2010-gg}
Kato Y, Matsubara S.
\newblock Correcting errors in a treebank based on synchronous tree substitution grammar.
\newblock In: Proceedings of the {ACL} 2010 Conference Short Papers. ACLShort '10. USA: Association for Computational Linguistics; 2010. p. 74--79.

\bibitem{Manning2011-bq}
Manning CD.
\newblock {Part-of-Speech} Tagging from 97\% to 100\%: Is It Time for Some Linguistics?
\newblock In: Computational Linguistics and Intelligent Text Processing. Springer Berlin Heidelberg; 2011. p. 171--189.

\bibitem{Nguyen2015-ul}
Nguyen PT, Le AC, Ho TB, Nguyen VH.
\newblock Vietnamese treebank construction and entropy-based error detection.
\newblock Language Resources and Evaluation. 2015 Sep;49(3):487--519.

\bibitem{Zeng2021-st}
Zeng Q, Yu M, Yu W, Jiang T, Jiang M.
\newblock Validating Label Consistency in {NER} Data Annotation.
\newblock In: Proceedings of the 2nd Workshop on Evaluation and Comparison of {NLP} Systems. Punta Cana, Dominican Republic: Association for Computational Linguistics; 2021. p. 11--15.

\bibitem{Chong2022-fa}
Chong D, Hong J, Manning CD.
\newblock Detecting Label Errors by using {Pre-Trained} Language Models.
\newblock arXiv. 2022 May.

\bibitem{Eskin2000-bn}
Eskin E.
\newblock Detecting Errors within a Corpus using Anomaly Detection.
\newblock In: 1st Meeting of the North {A}merican Chapter of the Association for Computational Linguistics; 2000. .

\bibitem{Nakagawa2002-tw}
Nakagawa T, Matsumoto Y.
\newblock Detecting Errors in Corpora Using Support Vector Machines.
\newblock In: {COLING} 2002: The 19th International Conference on Computational Linguistics; 2002. .

\bibitem{Dligach2011-df}
Dligach D, Palmer M.
\newblock Reducing the Need for Double Annotation.
\newblock In: Proceedings of the 5th Linguistic Annotation Workshop. Portland, Oregon, USA: Association for Computational Linguistics; 2011. p. 65--73.

\bibitem{Amiri2018-hj}
Amiri H, Miller T, Savova G.
\newblock Spotting Spurious Data with Neural Networks.
\newblock In: Proceedings of the 2018 Conference of the North {A}merican Chapter of the Association for Computational Linguistics: Human Language Technologies, Volume 1 (Long Papers). New Orleans, Louisiana: Association for Computational Linguistics; 2018. p. 2006--2016.

\bibitem{Swayamdipta2020-yr}
Swayamdipta S, Schwartz R, Lourie N, Wang Y, Hajishirzi H, Smith NA, Choi Y.
\newblock Dataset Cartography: Mapping and Diagnosing Datasets with Training Dynamics.
\newblock arXiv. 2020 Sep.

\bibitem{Yaghoub-Zadeh-Fard2019-dd}
Yaghoub-Zadeh-Fard MA, Benatallah B, Chai~Barukh M, Zamanirad S.
\newblock A Study of Incorrect Paraphrases in Crowdsourced User Utterances.
\newblock In: Proceedings of the 2019 Conference of the North {A}merican Chapter of the Association for Computational Linguistics: Human Language Technologies, Volume 1 (Long and Short Papers). Minneapolis, Minnesota: Association for Computational Linguistics; 2019. p. 295--306.

\bibitem{Wang2019-zf}
Wang Z, Shang J, Liu L, Lu L, Liu J, Han J.
\newblock {CrossWeigh}: Training named entity tagger from imperfect annotations.
\newblock In: Proceedings of the 2019 Conference on Empirical Methods in Natural Language Processing and the 9th International Joint Conference on Natural Language Processing ({EMNLP-IJCNLP}). Stroudsburg, PA, USA: Association for Computational Linguistics; 2019. .

\bibitem{Northcutt2021-ee}
Northcutt C, Jiang L, Chuang I.
\newblock Confident Learning: Estimating Uncertainty in Dataset Labels.
\newblock J Artif Intell Res. 2021 May;70:1373--1411.

\bibitem{Rehbein2017-ld}
Rehbein I, Ruppenhofer J.
\newblock Detecting annotation noise in automatically labelled data.
\newblock In: Proceedings of the 55th Annual Meeting of the Association for Computational Linguistics (Volume 1: Long Papers). Vancouver, Canada: Association for Computational Linguistics; 2017. p. 1160--1170.

\bibitem{Lee2020-id}
Lee J, Yoon W, Kim S, Kim D, Kim S, So CH, Kang J.
\newblock {BioBERT}: a pre-trained biomedical language representation model for biomedical text mining.
\newblock Bioinformatics. 2020 Feb;36(4):1234--1240.

\end{thebibliography}

\newpage
\appendix
\setcounter{table}{0}
\renewcommand\tablename{Supplementary Table}

\section{Supplementary materials}
\begin{table}[hbt!]
    \centering
    \vspace{1em}
    \small
    \caption{Definitions and examples of Family Relationship, Physical Health, and Mental Health Crises in the NVDRS coding manual.}
    \begin{tabular}{p{3cm}p{8cm}p{4cm}}
        \toprule
         Crisis & Definition & Example \\
         \midrule
         Family Relationship	& Code as `Yes' if at the time of the incident the victim was experiencing a relationship problem with a family member other than an intimate partner (e.g., a child, mother, in-law), and this appears to have contributed to the death. & Examples include the victim has an argument with his parents about a school suspension the night before the suicide. \\
         \midrule
         Physical Health	& Code as ‘Yes’ if the victim was experiencing physical health problems (e.g., terminal disease, debilitating condition, chronic pain) that were relevant to the event. Endorse this variable only if a health problem is noted as contributing to the death (e.g., despondent over recent diagnosis of cancer or complained that he could not live with the pain associated with a condition). Health conditions are coded from the perspective of the victim. If the victim believed him- or herself to be suffering from a physical health problem, and this belief was contributory to the death, it does not matter if any particular health problem was ever treated, diagnosed, or even existed. & Examples include the victim being despondent over being diagnosed with cancer two days before the suicide, a victim finding out the day before the suicide that their condition got significantly worse, or a victim just went bankrupt due to treatment for a chronic mental illness the day before the suicide.\\
         \midrule
         Mental Health & Code a person as `Yes' for if he or she has been identified as having a mental health problem. Mental health problems include disorders and syndromes listed in the DSM-5 (Diagnostic and Statistical Manual of Mental Disorders, 5th Revision) with the exception of alcohol and other substance disorders (as these are captured in separate variables). Also code `Yes' if the person was being treated for a mental health problem including treatment through involuntary mechanisms such as an Emergency Order of Detention, even if the nature of the problem is unclear (e.g., “was being treated for various psychiatric problems”). It is acceptable to endorse this variable on the basis of past treatment of a mental health problem, unless it is specifically noted that the problem has been resolved. Code `Yes' if a mental health problem is noted even if the timeframe is unclear (as in "history of depression"), or if the person was seeking mental health treatment or someone was seeking treatment on his or her behalf (e.g., "family was attempting to have him hospitalized for psychiatric problems"). This should also be coded as `Yes" if the IPV Victim or Perpetrator has a prescription for an antidepressant or other psychiatric medication. The drug list provided in the training notebook identifies drugs that can be considered psychiatric medications. We have separate questions for substance use problems. Therefore, do not include substance abuse as a "current mental health problem." & Examples of disorders qualifying as mental health problems include not only diagnoses such as major depression, schizophrenia, and generalized anxiety disorder, but developmental disorders (e.g., intellectual disability, autism, attention deficit hyperactivity disorder), eating disorders, personality disorders, and organic mental disorders such as Alzheimer's and other dementias.\\
         \bottomrule
    \end{tabular}
    \label{tab:supplement_table1}
\end{table}

\subsection{Crisis and state selection, training details}
\label{supplementaryA}
In our prior work, we applied BERT-based models to classify crises in NVDRS narratives 3. We selected Physical Health, Family Relationship and Mental Health crises in this study due to their higher frequency of positive instances and poor classification AUC scores compared to other crises (Table 2 and Figure 2 in Wang et al.~\cite{Wang2023-ov}). Similarly, we chose Ohio and Colorado as illustrative states for their higher frequency of positive instances and superior state-wise classification \fscore scores compared to other states (Table A2 and Table A3 in Wang et al.~\cite{Wang2023-ov}).
 
Binary Cross Entropy Loss and Adam optimizer were used during model training. We trained all the models for 30 epochs, and model selection was based on their performances on validation sets. The framework was implemented using PyTorch. We conducted our experiments using an Intel Xeon 6226R 16-core processor and Nvidia RTX A6000 GPUs.
\end{document}